\documentclass[10pt,twocolumn,letterpaper]{article}

 \usepackage{cvpr}              % To produce the CAMERA-READY version
\usepackage{multirow}
\usepackage{colortbl}
\usepackage{colortbl}
\usepackage{pifont}       % \ding{xx}
\usepackage{bbding}       % \Checkmark,\XSolid,... (需要和pifont宏包共同使用)
\usepackage{fontawesome}  % \faCheck,\faTimes

\definecolor{cvprblue}{rgb}{0.21,0.49,0.74}
\definecolor{mypink}{RGB}{255, 100, 203}
\usepackage[pagebackref,breaklinks,colorlinks,allcolors=cvprblue]{hyperref}

\def\paperID{11705} % *** Enter the Paper ID here
\def\confName{CVPR}
\def\confYear{2026}

\title{PC-Talk: Precise Facial Animation Control for Audio-Driven\\ Talking Face Generation}

\author{Baiqin Wang$^{1,2}$, Xiangyu Zhu$^{1,2}$\footnotemark[1], Fan Shen$^{3}$,Hao Xu$^{3,4}$, Zhen Lei$^{1,2,5,6}$\\
	$^{1}$MAIS, Institute of Automation, Chinese Academy of Sciences\\
	$^{2}$School of Artificial Intelligence, University of Chinese Academy of Sciences   \hspace{2mm}
	$^{3}$Psyche AI.INC \\ $^{4}$HKUST  \hspace{2mm}
	$^{5}$CAIR, HKISI, Chinese Academy of Sciences  \hspace{2mm}
	$^{6}$SCSE, FIE, M.U.S.T\\
	{\tt\small \{wangbaiqin2024, xiangyu.zhu, zhen.lei\}@ia.ac.cn}\\}
\begin{document}
\maketitle
{\renewcommand{\thefootnote}{\fnsymbol{footnote}}
	\footnotetext[1]{Corresponding author.}}
\begin{figure*}[t!]
	\centering
	\includegraphics[width=\textwidth]{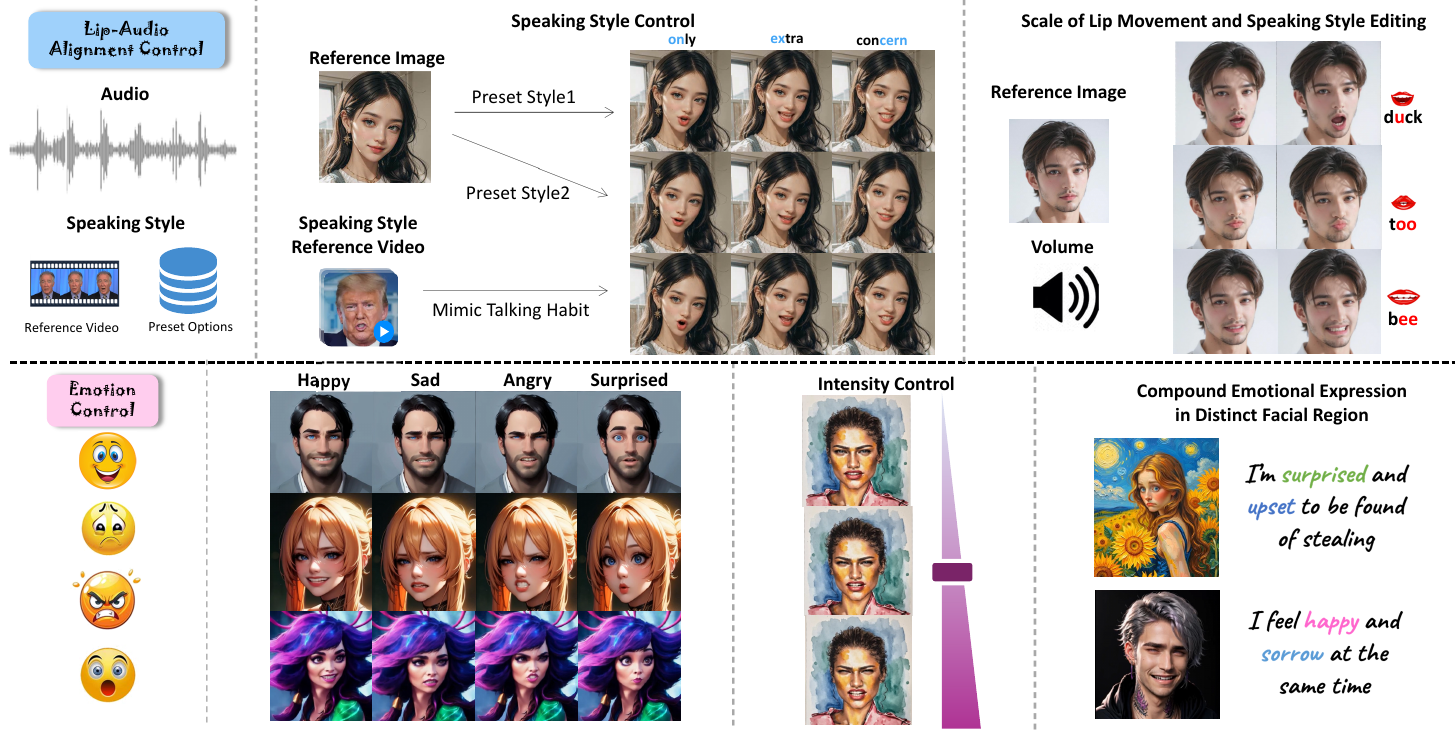}
	\vspace{-7mm} % 减少下方的垂直空白
	\caption{PC-Talk separates talking face control into two categories: Lip-Audio Alignment Control (LAC) for adapting and editing diverse speaking styles to simulate different talking habits, and EMotion Control (EMC) for generating expressive faces with adjustable intensity and region-specific compound emotions.  }
	\label{fig:first_pic}
	\vspace{-4mm} % 减少下方的垂直空白
\end{figure*}

\begin{abstract}{
		Recent advancements in audio-driven talking face generation have made great progress in lip synchronization. However, current methods often lack sufficient control over talking face, such as speaking style and emotional expression, resulting in uniform facial motion. In this paper, we focus on improving two key factors: lip-audio alignment control(LAC) and emotion control(EMC), to enhance the diversity and user-friendliness of talking videos. Lip-audio alignment control ensures accurate lip-sync across varied speaking styles to simulate different talking habits, whereas emotion control aims to generate realistic emotional expressions with varying intensities and mixed emotional states. To achieve precise facial animation control, we propose a novel and efficient framework, \textbf{PC-Talk}, which enables lip-audio alignment and emotion control through implicit keypoint deformations. First, our LAC module generates lip-synced talking faces with a specific speaking style, derived from either a video reference or preset options. It also supports lip movement scale adjustment and fine-grained editing of speaking styles for specific articulations. Second, our EMC module produces vivid emotional facial expressions through pure emotional deformation. It further enables precise control over emotion intensity and the compound emotions across different facial regions. Our method demonstrates outstanding control capabilities and achieves SOTA performance on HDTF and MEAD datasets in experiments. Project page:	 \textcolor{mypink}{https://bq-wang0511.github.io/PC-Talk/}
	}
\end{abstract}    
\section{Introduction}
\label{sec:intro}
Audio-driven talking face generation is increasingly prevalent in applications such as digital humans, film production, and voice assistants~\cite{zhao2025stavatar, guo2024liveportrait,wang20243d}. To meet the customization needs in these domains, it is crucial that the generated faces not only synchronize lip movements accurately with the corresponding audio but also offer users precise control for personalized generation. In this paper, we classify the controllability of talking face generation into two key aspects:  lip-audio alignment control and emotion control. 

Lip-audio alignment control focuses on achieving lip synchronization with diverse speaking styles to simulate different talking habits. Since each person has unique habits when pronouncing words, different speaking styles often result in subtle variations in lip articulation, even for the same phoneme. As illustrated in Fig.~\ref{fig:first_pic}, two distinct speaking styles produce noticeable differences in articulation, such as the openness of the mouth when saying "d\textbf{u}ck", the width of the mouth for "b\textbf{ee}", or the degree of lip pursing when pronouncing "t\textbf{oo}". We want to extract the speaking style from a reference video if available; otherwise, we can select from multiple preset options provided by the model, with the ability to edit each lip articulation individually. Additionally, controlling the lip movement scale helps simulate varying vocal loudness. On the other hand, emotion control focuses on generating expressive facial animations while maintaining lip-sync. As real human expressions often involve complex emotions beyond a single label, we aim to enhance the realism of emotional talking faces from two aspects. First, we want to support variation intensity for each emotion to enrich visual expressiveness and naturalness. Besides, since compound emotions are conveyed through different facial regions, such as a smiling mouth paired with a sorrowful eyebrow, it is essential to enable region-specific control for compound emotional expression. 

Previous methods have made significant progress in achieving accurate lip synchronization. For instance, Wav2Lip~\cite{prajwal2020lip} introduces a lip-sync discriminator to improve synchronization accuracy. Similarly, some approaches like~\cite{zhang2023sadtalker,tian2024emo,ji2025sonic,cui2024hallo2} can generate lip-synced videos from a single image. However, they lack robust control over the generated output. Some methods decompose the generation process into two stages: audio-to-motion and motion-to-image, leveraging intermediate representations to encode motion information. For example, EAT~\cite{gan2023efficient} utilizes implicit keypoints~\cite{wang2021one} as intermediate representations for emotional talking face generation, but falls short in the quality of the generated output. Recent approaches such as VASA~\cite{xu2025vasa} employ a DIT~\cite{peebles2023scalable} to produce lip-audio-synced motion, yet it has not fully exploited the potential for controllability within the motion space. In summary, achieving precise control remains a critical yet challenging goal in audio-driven talking face generation. 

% To address the challenges of controlling speaking style and emotional expression in talking face generation, we propose a novel framework based on implicit keypoint deformation. 
To address these challenges, we propose a novel and efficient talking face generation framework, \textbf{PC-Talk}, which enables precise facial animation control on several aspects such as speaking style and emotional expression. First, we propose a Lip-Audio alignment Control (LAC) module to achieve lip synchronization with diverse speaking styles. We first employ a style encoder to project both reference speaking style videos and preset style codes into a unified style space, enabling the model to support both types of style inputs during inference and thereby meeting the requirement for talking habit customization based on the input source. After that, we use a style-aware auto-regressive generator to produce talking face sequences conditioned on the desired speaking style. The scale of lip movements can be adjusted via a multiplicative factor applied to the output. To further edit speaking styles for different lip articulation, we introduce an edit module that modifies the motion projection of specific lip articulation. Second, we introduce an EMotion Control (EMC) module that can produce vivid emotional expressions. This module decomposes pure emotional deformations to synthesize highly expressive and emotionally distinct talking faces. Moreover, for compound emotional expressions, our approach enables the independent generation of emotions in specific facial regions, followed by seamless composition, leveraging the inherent bond between implicit keypoints and facial landmarks. Additionally, this module supports gaining emotional information from multiple optional sources, including explicit categorical labels, audio intonation, and the semantic content of the spoken text. 

In summary, our contributions are as follows:

\begin{enumerate}
	\item We propose a LAC module that generates lip-synced talking faces with diverse speaking styles to simulate different talking habit and supports further style editing. 
	\item We develop an EMC module that synthesizes vivid emotional expressions with controllable intensity and region-specific compound emotion from multiple sources. 
	\item Extensive experiments demonstrate that our framework not only provide precise control in talking face but also achieve state-of-the-art performance on  HDTF\cite{zhang2021flow} and MEAD\cite{wang2020mead} datasets.

\end{enumerate}

\section{Related Work}

\label{sec:related_work}

\subsection{Audio-driven Talking Face Generation}

Talking‑face generation aims to synthesize realistic facial animations with lip movements accurately synchronized to the input audio. Traditional methods typically adopt a one‑stage pipeline that directly generates lip‑synced frames from audio. Some approaches~\cite{cheng2022videoretalking,prajwal2020lip,zhang2024musetalk,li2024latentsync,wang2025talk} focus on predicting only the lower facial region to achieve lip synchronization—for instance, Wav2Lip~\cite{prajwal2020lip} employs a specialized lip‑sync expert network to improve alignment accuracy. More recent works~\cite{chen2024echomimic,cui2024hallo2,ji2025sonic} utilize generative models~\cite{peebles2023scalable,rogers2014diffusion} for full talking‑video synthesis, such as EchoMimic which builds upon Stable Diffusion~\cite{rombach2022high} and MultiTalk~\cite{kong2025let} which uses WAN 2.1~\cite{wan2025wan} as its backbone. While these approaches produce visually fluent results, they are computationally expensive and offer limited fine‑grained control over facial dynamics.

Recently, there has been growing interest in multi‑stage generation approaches. These methods~\cite{gan2023efficient,xu2025vasa,zhang2023sadtalker,li2025ditto} typically first predict an intermediate representation from audio and then reconstruct a lip‑synced image based on that representation. Among different facial representations, implicit keypoints have demonstrated strong expressive power. For example, Face‑vid2vid~\cite{wang2021one} achieves realistic animation by integrating 3D priors such as pose and expression deformations, while LivePortrait~\cite{guo2024liveportrait} establishes connections between implicit keypoints and explicit facial landmarks through landmark constraints. However, although several methods~\cite{gan2023efficient,xu2025vasa} adopt implicit keypoints for talking‑face generation, they overlook semantic consistency among keypoints and therefore require additional refinement. Ditto~\cite{li2025ditto} also employs implicit keypoints but suffers from unstable lip‑sync quality. In contrast, our method introduces semantically enriched lip‑related keypoints for accurate lip synchronization and incorporates a lip‑sync expert during training, inspired by Wav2Lip~\cite{prajwal2020lip}, achieving superior overall performance.

\subsection{Controllable Talking Face Generation}

Controllability in talking‑face generation is essential for enhancing user experience. For speaking‑style control, prior works~\cite{ma2023styletalk,zhang2023dream} typically extract a style code from a reference source and generate talking faces in the corresponding style. However, these approaches are often limited to a single style source and lack fine‑grained control, such as adjusting the shape or amplitude of specific lip articulations. For emotion control, previous studies~\cite{ji2022eamm,gan2023efficient,tan2024edtalk,tan2025disentangle} commonly generate facial expressions using emotion labels or reference images/videos. For instance, EAT~\cite{gan2023efficient} predicts emotional deformations to produce expressive talking faces but tends to suffer from lower image quality. ED‑Talk~\cite{tan2024edtalk} improves expressiveness by disentangling emotion from lip movements synchronized with audio, yet its control remains limited, lacking flexible adjustment of emotion intensity or compound emotional states due to constrained representations. Methods like DiceTalk\cite{tan2025disentangle} directly generate emotional talking faces from labels via large generative models, but still lack fine control and efficiency. In contrast, our method enables precise and continuous control over both speaking style and emotion, such as scaling lip‑movement range or modulating emotional intensity by manipulating the deformation of implicit keypoints.
\section{Method}
\label{sec:method}
\begin{figure*}[t]
	\centering
	\includegraphics[width=\textwidth]{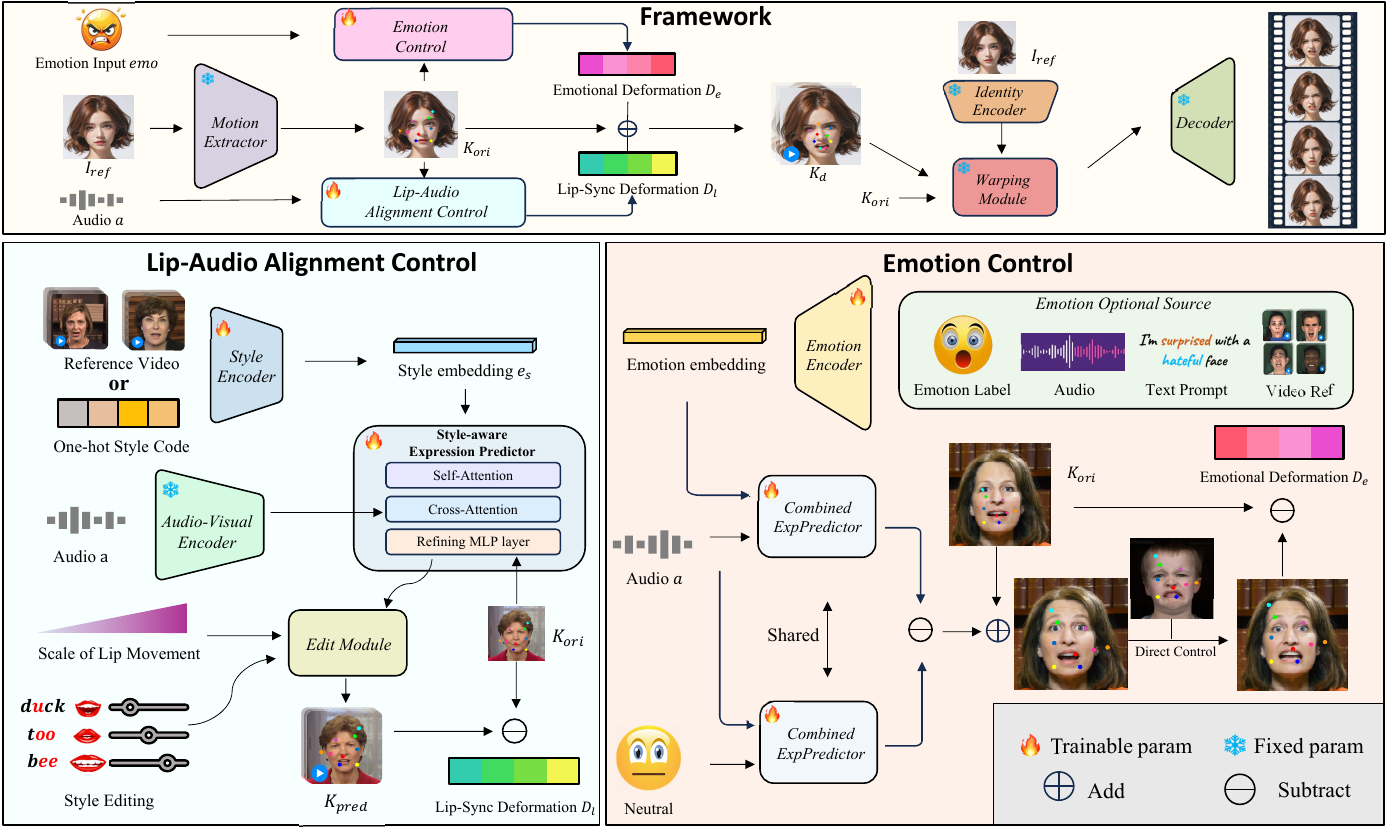}
	\vspace{-6mm}
	\caption{Our framework PC-Talk is designed for precise  control in talking face generation. It achieves this control by first predicting a deformation of implicit keypoints and then rendering it into a final talking image. We utilize a Lip-Audio alignment Control (LAC) module to estimate lip-sync deformations \(D_l\) and an EMotion Control (EMC) module to estimate emotional deformations \(D_e\). }
	\label{fig:overall}
	\vspace{-3mm}
\end{figure*}
\subsection{Preliminary}
As shown in Fig.~\ref{fig:overall},  our framework leverages implicit keypoints as an intermediate representation to achieve precise control over the talking face. The pipeline begins with a reference image \(I_{ref}\), from which we extract implicit keypoints \(K_{ori}\) using a motion extractor. Our motion extractor consists of three key components: a pose estimator for calculating rotation \(R\), transformation \(t\), and scale factor \(s\); an expression estimator for assessing expression deformation \( \delta  \); and a canonical keypoint detector for identifying the original keypoints \(K_c\) from various identity reference images. The computation of \(K_{ori}\) is conducted as follows: 
\begin{equation}
	K_{ori}=s \cdot\left(K_c \cdot R+\delta\right)+t
	\label{eq:transform}
\end{equation}

Note that some of these implicit keypoints possess semantic meaning, such as those corresponding to the lips or eyebrows, as they are tied to 2D facial landmarks by using a landmark distance constraint during training. Once the keypoints are extracted, our framework achieves controllable  talking face generation by calculating a lip-sync deformation \(D_l\) from LAC module and an emotional deformation \(D_e\) from the EMC module. These deformations are then combined to produce the driven keypoints \(K_{d} \) as follows:
\begin{equation}
	K_{d}= K_{ori}+D_l+D_e
	\label{eq:plus}
\end{equation}

We then render the final image \(I_{res}\) using a warping module and a decoder. The warping module estimates a flow field between \(K_{ori}\) and \(K_{d} \), which is applied to the appearance feature \(f_a\). Here, \(f_a\) is extracted from the reference image \(I_{ref}\) using an identity encoder. The decoder subsequently generates the final image \(I_{res}\) from the warped feature. The entire process can be summarized as follows: 
\begin{equation}
	I_{res}= \text{Decoder}\left(\text{Warp}\left( f_{a}, K_{ori}, K_{d} \right) \right)
	\label{eq:warp}
\end{equation}

During the training of the LAC and EMC modules, the parameters of other components are kept fixed, as the implicit keypoints serve as the facial representation.

\subsection{Lip-Audio Alignment Control}

\noindent\textbf{Style-aware Expression Predictor.} To achieve precise lip synchronization, we first employ an expression predictor to understand the general relationship between audio and lip movements. Inspired by FaceFormer~\cite{fan2022faceformer}, we utilize a style-aware auto-regressive Transformer~\cite{vaswani2017attention} to predict expression deformations.  The style embedding is added to the input of the model alongside the positional embedding, enabling style-conditioned generation. Additionally, the auto-regressive nature of the model promotes temporal consistency in the generated outcomes.  As for audio embedding \(e_a\), unlike existing methods that rely on pretrained ASR models such as Whisper~\cite{radford2023robust}, we pretrain the audio encoder on 2D audio-visual synchronization tasks, ensuring better alignment between audio features and lip movements.
The model consists of a self‑attention layer to fuse style and motion features, followed by a cross‑attention layer that interacts with audio features. Finally, a refinement MLP is applied to adjust the output directly on the implicit keypoints $K_{ori}$ rather than on expression deformations, mitigating the residual entanglement in the implicit representation. The expression predictor can be formulated as:

\begin{equation}
	\begin{aligned}
		D_l=\text{ExpPredictor}\left(e_a,e_s,K_{ori}\right),
	\end{aligned}
	\label{eq:expPred}
\end{equation}

\begin{table*}[t]
	\centering
	\caption{Quantitative comparisons with state-of-the-art methods.}
	\vspace{-3mm}
	\resizebox{\textwidth}{!}{
		\begin{tabular}{c|c|ccccc|ccccc}
			\toprule
			\multicolumn{1}{c}{\multirow{2}[4]{*}{\textbf{Method}}}  & 
			\multicolumn{1}{c}{\multirow{2}[4]{*}{\textbf{Input}}} & \multicolumn{5}{c}{\textbf{HDTF}} & \multicolumn{5}{c}{\textbf{MEAD-Neutral}}\\
			\cmidrule(lr){3-7}  \cmidrule(lr){8-12}  \multicolumn{1}{c}{} & \multicolumn{1}{c}{} &
			\multicolumn{1}{c}{LSE-C$\uparrow$}& \multicolumn{1}{c}{LSE-D$\downarrow$}  & \multicolumn{1}{c}{FID$\downarrow$}  & \multicolumn{1}{c}{$\text{NIQE}\downarrow$} & \multicolumn{1}{c}{$\text{FVD}\downarrow$}& \multicolumn{1}{c}{LSE-C$\uparrow$} & \multicolumn{1}{c}{LSE-D$\downarrow$}  & \multicolumn{1}{c}{FID$\downarrow$} & \multicolumn{1}{c}{$\text{NIQE}\downarrow$} & \multicolumn{1}{c}{$\text{FVD}\downarrow$}
			\\
			\midrule
			Wav2Lip \cite{prajwal2020lip} &Video& 8.65 & \underline{6.78} & 32.24 & \underline{13.82} & 183.99 & \underline{8.17} & \textbf{6.99}  & 54.40 & 13.28 & 393.06  \\
			VideoRetalking \cite{cheng2022videoretalking} &Video&  8.22 & 6.85 & 25.52 & 14.33& \underline{115.34} & 7.65 & \underline{7.22} & 41.86 & 13.38 & \underline{184.01}  \\
			MuseTalk \cite{zhang2024musetalk} &Video& 4.63 & 10.26  &  \underline{15.78} &14.22& 136.15 & 3.55 & 10.49 & 40.03& 12.89 & 249.03  \\
			LatentSync \cite{li2024latentsync} &Video&  \underline{8.92} & 6.84  &  16.32 &14.17& 175.23 & 8.15 & 7.67 &\underline{39.97} & \underline{12.76} & 287.05  \\
			\midrule
			\rowcolor{green!12}\textbf{PC-Talk}( Ours )  &Video& \textbf{9.03} & \textbf{6.69} & \textbf{15.51} & \textbf{13.49} &  \textbf{100.85}& \textbf{8.29} &7.59 & \textbf{24.88} & \textbf{12.04} & \textbf{170.56}  \\ 
			\midrule
			SadTalker \cite{zhang2023sadtalker} & Image & 7.15 & 7.93 & 40.75 & 46.30 &291.66& 6.36 & 8.24 & 63.70 & 32.59  & 193.84 \\
			Echomimic \cite{chen2024echomimic} & Image & 5.94 & 9.11  & \textbf{28.13} & 13.88 & 284.28 & 5.35 & 9.51 & \underline{40.92} & \underline{13.08} &216.72 \\
			Hallo-v2 \cite{cui2024hallo2}& Image & 7.53 & 7.97  & 33.15 & 13.80 & \underline{205.60} & 6.28 & 8.74  & 49.17 & 13.47 & 191.89 \\     
			Sonic \cite{ji2025sonic}& Image &\underline{8.64} & \underline{6.77}  & 40.81 & 14.16 & 212.78 & \underline{8.03} & \underline{7.34}  & 53.94 & 13.95 & \underline{173.54} \\     
			\midrule
			\rowcolor{green!12}\textbf{PC-Talk}( Ours ) &Image& \textbf{9.37} & \textbf{6.44} & \underline{33.07} & \textbf{13.29} & \textbf{205.55} & \textbf{8.19} &\textbf{7.69} &  \textbf{35.22} &  \textbf{12.48} &  \textbf{153.81}  \\ 
			
			\bottomrule
			
		\end{tabular}%
	}

	\label{tab:compare_all}%
\end{table*}%

\begin{table*}[htbp]
	\centering
	\caption{Quantitative comparisons on emotional talking face generation.}
	\vspace{-3mm}
	\resizebox{\textwidth}{!}{
		\begin{tabular}{c|cccccc|ccccccc}
			\toprule
			\multicolumn{1}{c}{\multirow{2}[4]{*}{\textbf{Method}}} & \multicolumn{6}{c}{\textbf{HDTF}} & \multicolumn{7}{c}{\textbf{MEAD}}\\
			\cmidrule(lr){2-7}  \cmidrule(lr){8-14}  \multicolumn{1}{c}{} & \multicolumn{1}{c}{LSE-C$\uparrow$}& \multicolumn{1}{c}{LSE-D$\downarrow$}  & \multicolumn{1}{c}{FID$\downarrow$}  & \multicolumn{1}{c}{$\text{NIQE}\downarrow$} & \multicolumn{1}{c}{$\text{FVD}\downarrow$}& \multicolumn{1}{c}{$\text{Acc}_{\text{emo}}\uparrow$} & \multicolumn{1}{c}{LSE-C$\uparrow$} & \multicolumn{1}{c}{LSE-D$\downarrow$}  & \multicolumn{1}{c}{FID$\downarrow$} & \multicolumn{1}{c}{$\text{NIQE}\downarrow$} & \multicolumn{1}{c}{$\text{FVD}\downarrow$}& \multicolumn{1}{c}{$\text{Acc}_{\text{emo}}\uparrow$}
			& \multicolumn{1}{c}{$\text{E-FID}\downarrow$}
			\\
			\midrule
			EAMM \cite{ji2022eamm} & 4.24 & 9.95 & 62.56 & \underline{30.48} & 475.61  &  25.21 & 4.33 & 9.530  & \underline{89.52} & \underline{25.09} & 359.13  & 36.78  & 2.63 \\
			EAT \cite{gan2023efficient} & \underline{7.65} & \underline{7.97} & 59.18 &  39.66 & 340.80 & 32.13 & 3.77 & 12.77  & 109.91 & 27.58 & 385.23 & \underline{68.21}  & 2.54 \\
			ED-Talk \cite{tan2024edtalk} & 7.21 & 8.11 & \underline{58.19} &  44.27 & \underline{305.75} & \underline{45.21} &\underline{7.75} & \underline{7.81} & 131.69 & 30.17  & \underline{210.39}  & 57.45  & \underline{2.10} \\
			\midrule
			\rowcolor{green!12}\textbf{PC-Talk}( Ours ) & \textbf{8.56} & \textbf{7.20} & \textbf{24.05} & \textbf{13.76} & \textbf{270.85} & \textbf{46.19}  & \textbf{7.83} & \textbf{7.74} & \textbf{35.26} & \textbf{12.05} & \textbf{190.05}  & \textbf{72.32} & \textbf {1.88} \\ 
			
			\bottomrule
			
		\end{tabular}%
	}
	\vspace{-2mm}

	\label{tab:compare_emo}%
\end{table*}%
\noindent\textbf{Style Space Modelize.} One of our goals for speaking style control is to adapt the speaking style from a reference video when provided by the user, or alternatively offer several preset speaking styles when no reference is available. To this end, we first represent each speaking style in the dataset using a one-hot code, which serves as a preset option for users. For video reference, we extract expression deformations and encode them into a shared style space using a Transformer encoder. Similarly, the one-hot style code is projected into the same style space via an MLP layer. During training, we adopt a mixed training strategy that randomly alternates between using the one-hot preset code and the reference video’s speaking style. This design enables flexible speaking style control via either reference video input or predefined style codes.

\noindent\textbf{Scale of Lip Movement.}  To simulate the effect of loudness on lip motion, we multiply the implicit keypoint deformation \(D_l\) with a scaling factor \(f\) based on the amplitude of the audio input. This enables our model to mimic lip movement variations with vocal volume changes, enhancing the realism and flexibility of generated talking faces. 

\noindent\textbf{Speaking Style Editing.} To achieve speaking style editing on particular lip articulation, our method employs a similar approach to the lip movement scaling described previously. Initially, we derive implicit keypoint deformation vectors for specific lip articulations such as lip pursing and mouth widening as shown in Fig.~\ref{fig:overall}. The generated implicit keypoint deformation \(D_l\) is then projected onto the targeted articulation vector, followed by applying a scaling factor to adjust the lip shape dynamically.  This technique enables precise, pronunciation-specific style editing, significantly enhancing the expressive capabilities of our method.

\subsection{Emotion Control}

\noindent\textbf{Decomposing Pure Emotional Deformation.} In our framework, an original idea to generate emotional talking faces is to predict the deformation of emotional faces directly. However,  the deformations in lip region consist of both lip-sync deformations and pure emotional deformations in emotional talking faces. These two types of deformations are inherently intertwined, making them difficult to disentangle.

To overcome these limitations, we decompose pure emotional deformation by subtracting neutral expression from emotional expression. First, we predict the combined deformations for both emotional and neutral expressions using the same audio input, ensuring consistent lip-sync deformation across them. Then, we subtract the neutral combined deformation from the emotional combined deformation to isolate the pure emotional deformation. The whole process can be formulated as:
\begin{equation}
	\begin{aligned}
		D_e=\text{CPred} (emo,e_a)-\text{CPred}(\text{'neutral'},e_a),
	\end{aligned}
	\label{eq:Cpred}
\end{equation}
where \(\text{CPred}\) represent combined expression predictor, \(emo\) means specific emotion category condition. This approach effectively isolates emotional expressions and ensures they are seamlessly integrated with synchronized lip movements. By leveraging the same architecture of expression predictor in the LAC module with combined expression predictor \(\text{CPred}\), we extend scale and style control to emotional expressions in EMC module, enabling applications such as modifying emotion intensity.

\noindent\textbf{Various Emotion Source.} Our method supports using multiple sources to flexibly control the emotional expressions of a talking face. For direct control sources such as images, we replace the original facial expression with the one extracted from the target image, enabling straightforward and effective emotion transfer. Conversely, for more complex control sources like audio or text, we derive emotional embeddings from these sources using pretrained emotion classifier networks. Further details are provided in the supplementary.

\noindent\textbf{Compound Emotional Expression Generation.} Complex emotions often combine multiple categories across facial regions, such as smiling with the mouth while expressing sadness through the eyes. Our method leverages implicit keypoint with semantic meanings in facial regions to generate emotional expressions for each facial region independently. These region-specific expressions are then seamlessly integrated to synthesize a talking face capable of capturing complex and nuanced emotional states. 

\subsection{Training and Inference}

\begin{figure*}[t!]
	\centering
		\includegraphics[width=\textwidth]{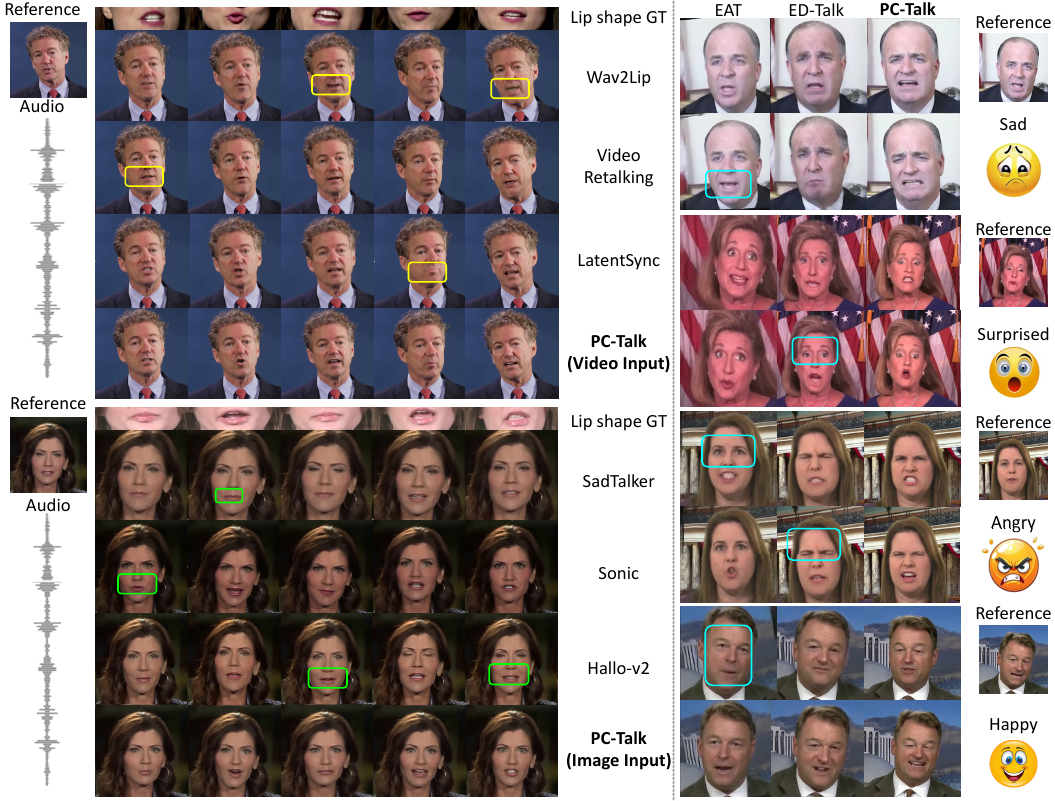}
	\vspace{-5mm}
	\caption{Comparison with other baselines. We highlight flaws of other methods using colorful bounding boxes, including blurry teeth, inaccurate lip shapes, and incorrect emotional expressions. Please zoom in to check details.}
	\label{fig:compare_all}
	\vspace{-3mm}
\end{figure*}

During the training process of LAC module, several kinds of loss are used as follows:

\begin{equation}
	\begin{aligned}
		\mathcal{L}_{LAC}=\mathcal{L}_{sync}+\lambda_{kp} \mathcal{L}_{kp}+\lambda_{reg} \mathcal{L}_{reg}\\+\lambda_{vel} \mathcal{L}_{vel}+\lambda_{style} \mathcal{L}_{style},
	\end{aligned}
	\label{eq:keypoint}
\end{equation}
where \(\mathcal{L}_{sync}\) is formulated as:
\begin{equation}
	\begin{aligned}
		\mathcal{L}_{s y n c}&=-\frac{\mathbf{S}_v\left(I_{gt: gt+4}\right)^{\mathbf{T}} \cdot \mathbf{S}_a\left(a_{gt: gt+4}\right)}{\left\|\mathbf{S}_v\left(I_{gt: gt+4}\right)\right\|_2\left\|\mathbf{S}_a\left(a_{gt: gt+4}\right)\right\|_2},
	\end{aligned}
	\label{eq:loss_sync}
\end{equation}
where \(I_{gt: gt+4}\) is a sequence of frames as image input, and  \(a_{gt: gt+4}\) is the  audio input. The sync loss $\mathcal{L}_{\text {sync }}$ is adapted from Wav2Lip~\cite{prajwal2020lip} , significantly enhances the model's ability to achieve accurate lip synchronization. The keypoint loss $\mathcal{L}_{kp}$ is L1 loss on implicit keypoint, and the regularize loss $\mathcal{L}_{reg}$ is employed to constrain excessive deformation changes, ensuring stable and natural results.   $\mathcal{L}_{\text {vel }}$ is used to enforce temporal consistency. $\mathcal{L}_{\text {style }}$ is a discriminative loss  to enhance the ability of style adapting. For EMC module, we only utilize  $\mathcal{L}_{kp}$, $\mathcal{L}_{reg}$ and  $\mathcal{L}_{\text {vel }}$. Please check the supplementary for details of loss functions.

Our method can seamlessly extend to video input by processing each frame as \(K_{ori}\). For video inputs, the pose source is derived from the video itself, while for image inputs, it is randomly selected from a set of predefined templates. During inference, we employ an overlap window to ensure temporal consistency in the auto-regressive model. Additionally, hyper-parameters such as the scaling factor can be adjusted, enabling the synthesis of realistic and customizable results tailored to specific requirements.

\section{Experiments}

\label{sec:experiments}
\subsection{Experimental Settings }
\textbf{Dataset.} We evaluate our model on the HDTF~\cite{zhang2021flow} and MEAD~\cite{wang2020mead} datasets. HDTF contains 16 hours of high-resolution videos with over 300 subjects, while MEAD includes 40+ identities with eight emotion types. The LAC module is trained on HDTF and neutral clips from MEAD, and the EMC module is trained on MEAD's emotional content. Training and test sets are split without overlap, with the test set containing 10-second audio clips for cross-identity inference. 

\noindent\textbf{Comparison Baselines.} We compare our method against three types of talking face generation approaches: video input methods including Wav2Lip~\cite{prajwal2020lip}, VideoRetalking~\cite{cheng2022videoretalking}, MuseTalk~\cite{zhang2024musetalk}, LatentSync~\cite{li2024latentsync},  single image input methods like SadTalker~\cite{zhang2023sadtalker}, Echomimic~\cite{chen2024echomimic}, Hallo-v2~\cite{cui2024hallo2}, Sonic~\cite{ji2025sonic} and emotional talking face generation methods such as EAMM~\cite{ji2022eamm}, EAT~\cite{gan2023efficient}, ED-Talk~\cite{tan2024edtalk}. This ensures a comprehensive evaluation across diverse scenarios. 

\noindent\textbf{Implementation Details.} We preprocess the dataset by converting videos to 25 fps and sampling audio at 16 kHz. The LAC and EMC modules are trained separately using the Adam optimizer~\cite{kingma2014adam} with a learning rate of \(1e\text{-}4\). Training is conducted on an RTX 4090 GPU, with the LAC module trained for two days and the EMC module for one day.  Notably, we use the same configuration for all comparisons without additional controls to adjust the final results. Our method produces outputs at a resolution of 512×512. The framework runs at 30 frames per second, supporting real-time generation. Additional implementation details are provided in the supplementary material. 

\noindent\textbf{Metrics.} We evaluate performance across three aspects: lip synchronization, image quality, and temporal consistency.  For lip sync, we use LSE-C and LSE-D from SyncNet~\cite{Chung_2016_syncnet}.  Image quality is assessed using the FID \cite{Seitzer2020FID} as a full-reference metric and NIQE~\cite{mittal2012making} for no-reference evaluation. Temporal consistency is measured via the FVD \cite{unterthiner2019fvd} .  Additionally, to assess emotional expressiveness, we adopt \(\text{Acc}_{\text{emo}}\) \cite{meng2019frame} for emotion classification and E-FID~\cite{deng2019accurate} for expression-distance evaluation. Since HDTF only contains neutral faces, \(\text{Acc}_{\text{emo}}\) is reported only on HDTF, while both metrics are used on MEAD. 

\begin{figure}[t!]
	\centering
	\includegraphics[width=\linewidth]{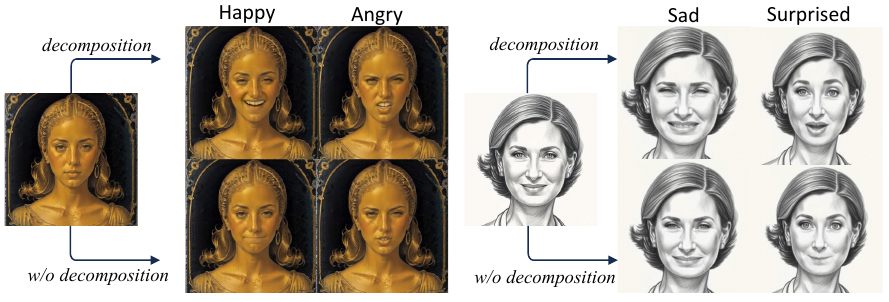}
	\vspace{-5mm}
	\caption{Ablation study on emotion decomposition.}
	\vspace{-3mm}
	\label{fig:ablation}
	
\end{figure}
\begin{figure}[t!]
	\centering
	\includegraphics[width=\linewidth]{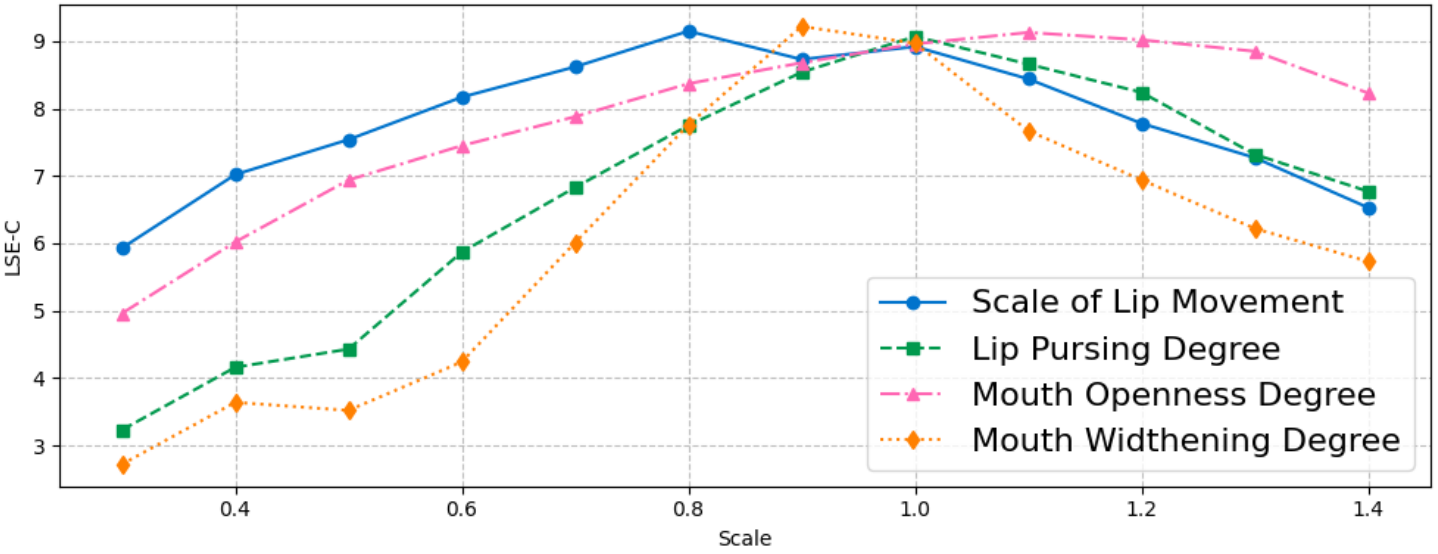}
	\vspace{-6mm}
	\caption{Lip-sync performance with different lip movement scales and speaking style editing across each lip articulation.}
	\vspace{-3mm}
	\label{fig:style_editing}
	
\end{figure}
\subsection{Evaluation on Lip-Audio Alignment}

\noindent\textbf{Quantitative Results.} As shown in Tab.~\ref{tab:compare_all}, our method achieves state-of-the-art performance across most evaluated dimensions in both HDTF and MEAD-Neutral datasets. For lip synchronization, our approach outperforms other baselines, achieving an LSE-C score of up to 9.03 with video input and 9.37 with image input in HDTF—a remarkable result that even surpasses Wav2Lip~\cite{prajwal2020lip} and LatentSync~\cite{li2024latentsync} which specifically designed for lip synchronization. In terms of image quality, our method also ranks high as the implicit keypoint renderer is well trained. As for temporal consistency, our method excels as well, thanks to the auto-regressive model and overlapping window during inference. 

\noindent\textbf{Qualitative Results.} As shown in Fig.~\ref{fig:compare_all}, our method demonstrates superior performance compared to other methods. In terms of lip synchronization, our approach achieves more precise lip shapes, whereas other methods frequently produce misaligned lip shapes compared to the ground truth. Regarding image quality, other methods suffer from teeth blurring and identity inconsistencies, while our method consistently delivers high-quality results. These findings underscore the effectiveness of our LAC module.

\subsection{Evaluation on Emotion}

\begin{table}[t!]
	\footnotesize
	\caption{Ablation Study on lip-audio alignment.} 
	\vspace{-6mm}
	\begin{center}
		{
			
			\begin{tabular}{c|cccc}
				\toprule
				AV Encoder &{\ding{55}} &{\ding{51}} & {\ding{51}} & {\ding{51}} \\
				$\mathcal{L}_{\text {sync }}$ & {\ding{55}} & {\ding{55}} & {\ding{51}}  &{\ding{51}}\\
				$\mathcal{L}_{\text {kp }}$ & {\ding{55}} & {\ding{55}} & {\ding{55}}  & {\ding{51}}\\ \midrule
				LSE-C  $\left(\uparrow\right)$ &6.23 & 7.17 & 8.92 & \textbf{9.37} \\
				
				\bottomrule
			\end{tabular}
		}
	\end{center}
	\vspace{-3mm}
	\label{tab:ablation}
\end{table}

\begin{table}[t]
	\scriptsize
	\centering 
		
	\caption{Efficiency Comparison (frame per second). }
	\vspace{-3mm}
	\resizebox{\columnwidth}{!}{%
		\begin{tabular}{cccccc}
			\toprule
			SadTalker  & EchoMimic 
			&  Hallo-v2 & Ours(\textit{w/o} control) & \textbf{Ours} \\
			\midrule
			10.76 & 0.84 & 0.69 & 34.75 & 30.13\\
			\midrule
		\end{tabular}%
	}
	\label{tab:eff}
	\vspace{-2mm}
\end{table}

\noindent\textbf{Quantitative Results.} As shown in Tab.~\ref{tab:compare_emo}, our method demonstrate superior performance in lip synchronization and  image quality compared to other emotional talking face methods.  Regarding emotion evaluation, our method attains the highest scores, with an \(\text{Acc}_{\text{emo}}\)  of 46.19 on the HDTF dataset and 72.32 on the MEAD dataset, , indicating that our method conveys emotions more accurately. Our method also surpasses others in E-FID on the MEAD dataset, further demonstrating the effectiveness of EMC module. 

\noindent\textbf{Qualitative Results.} As shown in Fig.~\ref{fig:compare_all}, our method achieves superior results compared to others. Our approach conveys emotions more effectively and vividly, with a significantly greater intensity. For instance, our method easily captures features such as narrowed eyes in anger, while EAT~\cite{gan2023efficient} fails to synthesize due to its limited expressive capability. Similarly, ED-Talk~\cite{tan2024edtalk} inaccurately blends expressions of anger, introducing elements of sadness into it due to its constrained emotion bank. The emotion categories are more clearly distinguishable in our results, reflecting the robustness of the EMC module. 

\subsection{Ablation Study}

\noindent\textbf{Lip-Audio Alignment.} As shown in Tab.~\ref{tab:ablation}, we conducted experiments on attributes related to the lip-audio alignment component, audio-visual encoder , sync loss $\mathcal{L}_{\text {sync }}$ and keypoint loss $\mathcal{L}_{kp}$. A noticeable drop in lip-sync performance is observed when replacing our audio-visual encoder with the conventional Whisper\cite{radford2023robust}. Moreover, the introduction of  $\mathcal{L}_{\text {sync }}$ leads to a notable improvement in LSE-C, and the addition of $\mathcal{L}_{kp}$ yields further enhancement.

\noindent\textbf{Emotion Decomposition.} As observed in Fig.~\ref{fig:ablation}, the  decomposition significantly enhances the expression of emotions. Without decomposition, facial emotions appear unnatural such as mouth over closed, while our method effectively captures and conveys emotions. This demonstrates a substantial improvement in the emotional expressiveness of decomposition of pure emotional deformation. 

\begin{figure}[t!]
%	\vspace{-2mm}
	\includegraphics[width=\linewidth]{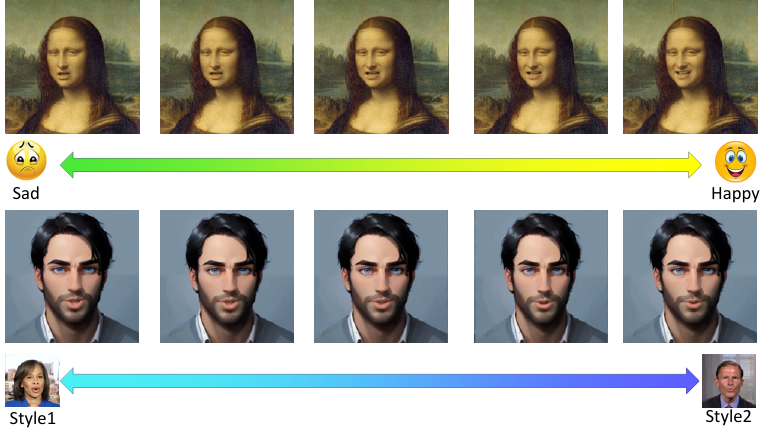}
	\vspace{-9mm}
	\caption{Emotion and Speaking Style Interpolation }
	\label{fig:inp}
\end{figure}

\subsection{Analysis}

\noindent\textbf{Scale of Lip Movement and Speaking Style Editing.} As shown in Fig.~\ref{fig:style_editing}, we conduct experiments on varying the scale of lip movement and the degree of style editing for each lip articulation. We observe that the optimal performance for lip movement scale occurs around 0.8. For style editing degrees, lip-sync performance remains stable between 0.8 and 1.2, but obviously degrades outside this range. Among different articulation types, mouth openness is less affected by scale changes, while lip pursing and mouth widthening are more sensitive. 

\noindent\textbf{Emotion and Speaking Style Interpolation.} As shown in Fig.~\ref{fig:inp}, our method supports both emotion and speaking style interpolation. For emotion interpolation, the facial expression gradually transitions from happy to sad, enabling applications such as dynamic emotion control. For speaking style interpolation, we interpolate style embeddings between the learned style space. The results demonstrate a smooth and continuous change in lip articulation between the two styles, confirming the consistency and effectiveness of the style space we constructed.

\noindent\textbf{Efficiency.} As demonstrated in Tab.~\ref{tab:eff}, even though we have integrated several control modules, our method exhibits only a slight degradation compared to the original animation framework. It also shows significant advantages over other methods including advanced diffusion methods like Hallo-v2~\cite{cui2024hallo2} and EchoMimic~\cite{chen2024echomimic}. 

\begin{table}[t!]
	\centering
	\caption{User Study results. The rating is on scale of 1-5, with the higher rank demonstrate better results.}
	\vspace{-2mm}
	\resizebox{\linewidth}{!}{
		
		\begin{tabular}{c|ccccc}
			\toprule
			%	Method & Lip Synchronization & Image quality & Temporal Consistency & Emotional Expressive & Overall realistic \\
			\multirow{2}{*}{Methods}  & Lip-sync  & Image  & Temporal & Emotional & Overall \\ 
			& Accuracy & Quality & Consistency & Expressive & Realistic\\
			\midrule
			Wav2Lip \cite{prajwal2020lip} & 3.3 & 1.9 & 1.9 & ----& 1.9 \\
			MuseTalk \cite{zhang2024musetalk} & 2.2 & 2.6 & 3.0  & ----& 3.4\\
			Echomimic \cite{chen2024echomimic}& 1.9 & 3.4  & 3.5  &----& 3.0\\ 
			Hallo-v2  \cite{cui2024hallo2}&3.8 & 4.0 & 3.4 & ---- & 3.8\\
			EAT  \cite{gan2023efficient}&1.3 & 2.9 & 2.6 & 2.6& 2.1 \\
			ED-Talk  \cite{tan2024edtalk}&3.1 & 1.4 & 2.4 & 2.3 &1.6\\ \midrule[1pt]
			\textbf{PC-Talk}(Ours)   &\textbf{4.6} & \textbf{4.8} & \textbf{4.6} & \textbf{4.9} &\textbf{4.8} \\
			
			\bottomrule
		\end{tabular}
	}
	\vspace{-5mm}

	\label{tab:user_study}%
\end{table}%

\subsection{User Study}
We conducted a user study evaluating four key dimensions: lip synchronization, image quality, temporal consistency, and emotional expressiveness. Additionally, we assessed the overall realism of the generated videos. For the evaluation, we used 10 generated video clips from each method, each lasting more than 5 seconds. The questionnaire followed the Mean Opinion Score protocol and involved more than 20 participants.  As presented in Tab.~\ref{tab:user_study}, our method consistently achieved the highest scores across all evaluated dimensions compared to competing approaches. Moreover, when focusing specifically on methods for emotional talking face generation, our approach demonstrated particularly strong performance in emotional expressiveness. These results highlight the effectiveness of our method in producing talking faces that are not only visually realistic and temporally coherent but also emotionally rich.
\section{Conclusion}
\label{sec:conclusion}
We present PC-Talk, a framework that provides precise control over facial animation, including speaking style and emotional expression in audio-driven talking face generation. Our approach  consists of two key components:  Lip-Audio alignment Control (LAC) module and EMotion Control (EMC) module. LAC module produces lip-synced talking face videos with diverse speaking styles. EMC module generates vivid emotional expression from various sources, enabling emotional intensity modification and complex emotional expression synthesis. Our approach achieves state-of-the-art results on the HDTF and MEAD datasets and demonstrates exceptional controllable capabilities, greatly enhancing the user experience in talking face generation.

\section*{Acknowledgments}
\label{sec:acknowledge}
This work was supported in part by Beijing Natural Science Foundation L242092, Chinese National Natural Science Foundation Projects 92570119, 62276254, U23B2054,  the Science and Technology Development Fund of Macau Project 0140/2024/AGJ, and InnoHK program.
{
    \small
    \bibliographystyle{ieeenat_fullname}
    \bibliography{main}
}


\end{document}